\documentclass{article}
\usepackage{graphicx} % Required for inserting images
\usepackage{longtable}
\usepackage{amssymb}

\usepackage{comment}
\usepackage{adjustbox}

\usepackage{booktabs}
\usepackage{amsmath}
\usepackage{soul}

\usepackage[letterpaper, margin=1in]{geometry}
\usepackage[table]{xcolor}
\usepackage{caption}
\usepackage{booktabs}
\usepackage{tabularx}
\usepackage{tikz}
\usetikzlibrary{calc}
\usetikzlibrary{positioning, shapes, arrows.meta}
\usepackage{array}
\usepackage{longtable}

\definecolor{LowColor}{HTML}{90EE90}    % Light Green (L=1)
\definecolor{MediumColor}{HTML}{FFD700} % Gold/Yellow (M=2)
\definecolor{HighColor}{HTML}{FF8C00}   % Dark Orange (H=3)
\definecolor{CriticalColor}{HTML}{FF4500}% Orange Red (C=4)

\newcommand{\Rate}[1]{%
    \ifnum #1 = 1
        \cellcolor{LowColor!70}L (1)%
    \else\ifnum #1 = 2
        \cellcolor{MediumColor!70}M (2)%
    \else\ifnum #1 = 3
        \cellcolor{HighColor!70}H (3)%
    \else\ifnum #1 = 4
        \cellcolor{CriticalColor!70}C (4)%
    \fi\fi\fi\fi
}

\title{Security Risks of Agentic Vehicles: A Systematic Analysis of Cognitive and Cross-Layer Threats}
\author{Ali Eslami and Jiangbo Yu}
\date{}

\begin{document}
%\tableofcontents

\maketitle
	\begin{abstract}    
        Agentic AI is increasingly being explored and introduced in both manually driven and autonomous vehicles, leading to the notion of Agentic Vehicles (AgVs), with capabilities such as memory-based personalization, goal interpretation, strategic reasoning, and tool-mediated assistance. While frameworks such as the OWASP Agentic AI Security Risks highlight vulnerabilities in reasoning-driven AI systems, they are not designed for safety-critical cyber-physical platforms such as vehicles, nor do they account for interactions with other layers such as perception, communication, and control layers. This paper investigates security threats in AgVs, including OWASP-style risks and cyber-attacks from other layers affecting the agentic layer. By introducing a role-based architecture for agentic vehicles, consisting of a Personal Agent and a Driving Strategy Agent, we will investigate vulnerabilities in both agentic AI layer and cross-layer risks, including risks originating from upstream layers (e.g., perception layer, control layer, etc.). A severity matrix and attack-chain analysis illustrate how small distortions can escalate into misaligned or unsafe behavior in both human-driven and autonomous vehicles. The resulting framework provides the first structured foundation for analyzing security risks of agentic AI in both current and emerging vehicle platforms.
    \end{abstract}

\section{Introduction}
\subsection{Motivation}

Connected and autonomous vehicles (CAVs) increasingly integrate agentic AI
components that perform reasoning, maintain long-term memory, interpret user
intent, invoke external tools, and participate in multi-agent coordination \cite{polamarasetti2025agentic,zhu2025semagent,khamis2025agentic}.
These capabilities differ fundamentally from the deterministic control and
planning architectures that have traditionally governed automotive autonomy. As
vehicles begin to rely on AI-driven roles that shape goals, interpret context,
and justify behavioral choices, the attack surface expands beyond the familiar
domains of communication channels, perception pipelines, and embedded
electronics.

Agentic reasoning introduces vulnerabilities that differ from traditional cyberattacks.
Cognitive functions such as memory, goal management, and justification can be subtly manipulated, allowing small distortions in user intent or stored context to reshape vehicle behavior without triggering standard detectors \cite{sapkota2025ai}. These errors can propagate across agent roles, forming attack chains that bypass existing safety mechanisms. Because agentic reasoning depends on abstracted summaries from perception, communication, and computation layers, upstream attacks can also mislead the agentic pipeline even when the agents themselves are not directly compromised.

Despite the rapid deployment of foundational models and agentic AI in mobility
systems, the cybersecurity community lacks a systematic framework for analyzing
risks that arise from these cognitive behaviors. Existing automotive standards
address communication and hardware vectors \cite{ward2021automotive}, while emerging AI threat models
focus on generic LLM applications rather than safety-critical autonomy \cite{owasp2023owasp}. A
domain-specific analysis is needed to capture how agentic reasoning interacts
with vehicle architectures, how these interactions create novel vulnerabilities,
and how adversaries can exploit cross-role propagation to induce misaligned or
unsafe driving behavior.

Furthermore, while using agentic AI to detect cyber-attacks is receiving researchers attention recently \cite{polamarasetti2025agentic, gosmar2025sentinel}, there is an assumption on safety and security of the designed agents. Moreover, one needs to analyze the risks of agentic AI and discuss possible cybersecurity design implications. Therefore, These trends motivate the need for a principled analysis that explains what are agentic vulnerabilities, how they propagate across roles in a vehicle, and how
their impact differs from conventional CAV threats.

\subsection{Gap in Existing Literature}

Existing research on vehicle cybersecurity predominantly centers on the
traditional CAV stack: in-vehicle networks, V2X communication, perception
integrity, and control-system protection \cite{kifor2024automotive,abreu2025cybersecurity}. Standards such as ISO/SAE~21434 \cite{ward2021automotive} and
UNECE WP.~29 \cite{costantino2022comparative} provide structured methodologies for identifying hardware and
communication vulnerabilities, but they assume deterministic system components
with well-defined behaviors. A few works have recently considered safety and security of AI systems in vehicles \cite{ullrich2024ai}. However, these frameworks do not account for reasoning-driven
failures, memory-dependent misalignment, or goal manipulation—phenomena that
emerge only when vehicles incorporate agentic AI.

Recent AI-security works such as \cite{owasp2023owasp} identifies LLM risks such as prompt injection, tool misuse, and hallucinations, but these frameworks focus on chatbots and enterprise systems, not safety-critical cyber-physical platforms such as CAVs. They do not model how agentic reasoning influences driving strategies, how corrupted cognitive processes affect control decisions, or how cross-layer information flows amplify subtle manipulations. No existing framework unifies automotive cybersecurity with agentic AI behavior, leaving a gap in understanding how agentic roles interact with the CAV stack and how cognitive vulnerabilities propagate into safety-relevant outcomes.

This work addresses that gap by providing the first structured analysis of
agentic AI vulnerabilities in vehicles, grounded in a role-based architectural
model and integrated with cross-layer CAV considerations.

\subsection{Contributions}

This work makes four primary contributions toward understanding the security
implications of agentic AI in both manually driven and autonomous vehicles:

\begin{enumerate}
    \item We propose a principled decomposition of the agentic layer into the Personal
    Agent and Driving Strategy Agent. These agents can be considered as meta agents with possible sub-agents included in each one of them. The architecture
    clarifies responsibilities, trust boundaries, and authority levels, providing
    a foundation for systematic security analysis.

    \item
    We introduce a structured classification of threats that arise from cognitive
    behaviors such as memory formation, intent interpretation, tool invocation,
    cross-role reasoning, and long-horizon planning. The taxonomy captures both
    localized vulnerabilities and multi-stage attack pathways.

    \item
    We assess how each threat affects the three agentic roles and identify
    pathways through which localized distortions escalate into misaligned or
    unsafe driving behavior. The matrix highlights where mitigations must be
    prioritized, and then, we provide some example attack-chains and provide analysis on how they propagate through the system.

    \item
    We show how adversarial manipulation of perception, communication,
    and control layers can influence agentic reasoning indirectly,
    even without compromising the agents themselves. This integration bridges
    conventional CAV security with emerging AI-driven vulnerabilities.
\end{enumerate}

Together, these contributions provide the first structured framework for
analyzing the security of agentic AI in autonomous and semi-autonomous vehicles,
laying the groundwork for principled defense strategies and future research.

\subsection{Paper Organization}

The remainder of the paper is structured as follows. Section \ref{sec:Background & Related Works} reviews existing
automotive cybersecurity frameworks and prior works. Section \ref{sec:Agentic Vehicle Architecture} introduces the proposed role-based agentic
vehicle architecture and analyzes trust boundaries and inter-role dependencies.
Section \ref{sec:threat_taxonomy} presents the threat taxonomy, along with severity assessments and representative attack chains.
Section \ref{sec:Integrated_Attacks} examines how attacks originating in the perception, communication, and control layers propagate into the agentic pipeline and Section \ref{sec:Discussion} provides discussions and cybersecurity implications for vehicle design. Section \ref{sec:Case Studies}
provides some case studies and finally, Section \ref{sec:Conclusion} concludes the paper with key insights and avenues for
future work.

\section{Background \& Related Works}
\label{sec:Background & Related Works}
\subsection{Cybersecurity risks in Connected and Autonomous Vehicles: Layered Analysis}

Connected and Autonomous Vehicles (CAVs) rely on a tightly integrated pipeline that spans perception, in-vehicle control, communication, and cloud services \cite{sun2021survey}. Understanding cybersecurity risks in such systems requires analyzing how vulnerabilities originate in each layer and how attacks can propagate through the entire stack. Although each layer has its own technical role, adversarial manipulation at any point can distort the vehicle’s understanding of the environment, disrupt internal coordination, or mislead high-level decision modules, ultimately influencing safety-critical behavior.

The perception layer forms the vehicle's primary interface with the physical world, using sensors such as cameras, LiDAR, radar, ultrasonic units, GPS receivers, and inertial measurement sensors to generate raw or fused representations of the surrounding environment. Because these measurements depend directly on external physical signals, attackers can manipulate sensor inputs without breaching internal systems \cite{mahima2024toward}. Physically altered traffic signs, spoofed LiDAR returns, and forged GPS signals can induce a coherent but incorrect environmental understanding. Such distortions can cause an autonomous vehicle to misinterpret road conditions or lead the driver of a manually driven vehicle to receive misleading guidance.

Below perception lies the in-vehicle control and network layer, which includes Electronic Control Units (ECUs), gateway modules, and communication buses such as CAN, LIN, FlexRay, and automotive Ethernet \cite{anwar2023security}. This layer is responsible for translating control decisions into physical actuation, making it one of the most safety-critical components of the CAV stack. Adversaries may inject forged CAN messages, modify ECU firmware, or launch denial-of-service attacks on internal networks, preventing timely transmission of essential commands. While perception attacks mislead interpretation, compromises at this layer can directly change braking, steering, or propulsion behavior \cite{luo2023vehicle}.

The communication layer responsible for V2X connectivity introduces another category of risks \cite{sedar2023comprehensive}. Vehicles communicate with other vehicles, roadside infrastructure, and external devices using DSRC, C-V2X, Wi-Fi, Bluetooth, and cellular modems. Because these channels depend on wireless broadcasts and implicit trust relationships, attackers can impersonate legitimate actors, replay outdated information, or inject fabricated warnings and congestion reports \cite{sousa2017new}. Through such manipulation, they can influence not just a single vehicle but potentially entire regions of traffic, illustrating how V2X compromises amplify their impact through network effects \cite{lu2018survey}.

Finally, the cloud and application layer expands the attack surface beyond the vehicle itself \cite{li2024over}. Modern platforms depend on over-the-air update services, cloud-hosted models, map servers, and large-scale data aggregation \cite{gligorijevic2025security}. Compromise at this level can propagate globally: a malicious firmware update may reach an entire fleet, a poisoned training dataset may degrade perception performance system-wide, and corrupted map data may cause widespread routing failures. Human-vehicle interfaces—such as infotainment systems and companion apps—further extend this layer’s vulnerabilities. If compromised, they may reveal vehicle state, enable unauthorized tracking, or even provide a bridge into in-vehicle networks, despite appearing benign or user-facing.

Together, these layers reveal that cybersecurity in CAVs is inherently cross-domain. Attacks need not target deep software vulnerabilities; they can arise from physical manipulation, wireless impersonation, internal network tampering, or cloud-service compromise. As vehicles grow more software-defined and interconnected, vulnerabilities in any layer can mislead higher-level reasoning modules or distort physical behavior, underscoring the need for holistic, multi-layer security strategies in both current and emerging vehicle architectures.

\subsection{Agentic AI and Levels of Agency}

Agentic AI systems in vehicles differ from traditional perception or control
modules because they maintain internal state, pursue goals, coordinate tasks,
and interact with users or tools through reasoning and adaptation \cite{acharya2025agentic}. To clarify
the types of agentic behaviour relevant for threat modeling, we adopt a
six-level capability taxonomy consistent with contemporary agency frameworks \cite{yu2025agentic}:

\begin{itemize}
    \item \textbf{Level 0 – Reactive Tools:} 
    Purely rule-based systems with no memory, planning, or reasoning. 
    Examples include fixed ADAS alerts or deterministic rule triggers.

    \item \textbf{Level 1 – Assisted Agents:} 
    Systems that interpret context and provide recommendations or filtered
    information with short-term state, but cannot autonomously plan or use tools.

    \item \textbf{Level 2 – Task Agents:}
    Agents capable of bounded multi-step reasoning and executing specific tasks
    under user-defined goals. They maintain task-scoped memory and may access
    limited external tools or APIs.

    \item \textbf{Level 3 – Contextual Agents:}
    Agents that integrate long-horizon context, adapt objectives based on
    environmental conditions, and coordinate multiple actions or tasks. They rely
    on richer memory and can reason about trade-offs among goals.

    \item \textbf{Level 4 – Collaborative Agents:}
    Agents that coordinate with other agents, negotiate goals, share state, and
    maintain persistent role-specific memory. They can manage multi-agent task
    execution and resolve conflicts through reasoning.

    \item \textbf{Level 5 – General Agents:}
    Highly autonomous systems capable of dynamic goal-setting, abstract
    reasoning, cross-domain tool use, and reflective self-modification of plans
    or objectives. These represent the highest degree of agency and the broadest
    cognitive attack surface.
\end{itemize}

These six levels are \emph{orthogonal} to SAE levels of driving automation.
A vehicle might host AI agents corresponding to any agency level, depending on the type of the vehicle and the possibility of certification.

An important consideration is that higher levels of agency introduce progressively broader attack surfaces:
memory poisoning, intent manipulation, cascading hallucinations, and reasoning
misalignment become possible only when systems maintain state, plan tasks, or
interact autonomously with tools or other agents. This motivates the need for a
role-specific threat model for agentic vehicles, developed in the following
sections.

\subsection{Existing Threat Models and OWASP Agentic AI Risks}
Traditional automotive cybersecurity research has developed extensive 
frameworks for analyzing and mitigating risks in Connected and Autonomous 
Vehicles (CAVs) \cite{kifor2024automotive,abreu2025cybersecurity,luo2022cybersecurity}. However, these frameworks primarily focus on network 
security, in-vehicle communication authenticity, and physical-layer safety 
mechanisms. The emergence of \emph{Agentic AI} introduces reasoning-driven, 
memory-based, and goal-oriented behaviors that significantly expand the 
attack surface beyond the scope of existing automotive standards and threat 
models. This subsection reviews relevant prior work and clarifies the gap 
that motivates the present study.

\subsubsection{Traditional Automotive Security Standards}

Traditional automotive cybersecurity frameworks, such as ISO/SAE 21434 \cite{ferdous2025addressing} and UNECE WP.29 \cite{costantino2022comparative}, focus on hardware interfaces, in-vehicle networks, diagnostic ports, and telematics security. Their threat models assume deterministic components with predictable behaviors and do not address vulnerabilities arising from reasoning-driven systems with memory, goal adaptation, or tool invocation. As a result, these standards provide strong coverage for classical CAV attack surfaces but lack mechanisms for analyzing failures emerging from agentic cognition, long-term internal state, or semantic decision pathways.

\subsubsection{OWASP Agentic AI Security Risks}

The OWASP Agentic AI Security Risks highlight threats specific to autonomous, tool-using, memory-maintaining agents—such as memory poisoning, cascading reasoning errors, privilege misuse, and unsafe tool invocation \cite{owaspAgenticAI}. These risks arise from properties unique to agentic systems: persistent state, multi-step reasoning loops, dynamic objectives, and inter-agent communication. While the taxonomy is valuable, it is domain-agnostic and does not consider the real-time, safety-critical constraints of vehicles or the tight coupling between cognitive reasoning and the cyber-physical stack. Consequently, it does not capture how these risks manifest when agent outputs influence driving decisions.

\subsubsection{Gap in Existing Frameworks and Position of This Work}

Traditional automotive cybersecurity frameworks address hardware integrity, in-vehicle networks, and communication security, but they assume deterministic modules with fixed behaviors. They do not account for vulnerabilities stemming from agentic cognition, such as memory-driven misalignment, goal manipulation, or multi-step reasoning drift. Conversely, AI-centric taxonomies like the OWASP agentic risks capture cognitive threats—memory poisoning, cascading hallucinations, unsafe tool use—but remain domain-agnostic and do not incorporate the real-time safety constraints or hierarchical authority flows present in vehicles.

This leaves a gap at the intersection of agentic reasoning and cyber-physical operation. Current frameworks lack a unified view of how distorted intent, corrupted inter-agent messages, or upstream CAV-layer attacks propagate through a vehicle’s reasoning pipeline and influence physical behavior. This work addresses that gap by contextualizing the fifteen OWASP agentic risks within a role-based vehicular architecture and demonstrating how they interact with perception, V2X, fusion, and control layers to produce safety-critical consequences.

\section{Agentic Vehicle Architecture}
\label{sec:Agentic Vehicle Architecture}
\subsection{Role-Based Multi-Agent Design Principles}

The proposed agentic vehicle architecture adopts a role-based design in which
each cognitive function is encapsulated within a distinct agent with clearly
defined responsibilities, authorities, and communication interfaces. This
structure is motivated by the need to prevent uncontrolled capability
amplification: as agentic systems gain autonomy, unconstrained reasoning,
persistent memory, and tool invocation can create broad attack surfaces unless
their influence on the driving stack is strictly partitioned. A role-based
model therefore enforces the principle that no single agent can independently
interpret user intent, generate vehicle behavior, and validate safety. Instead,
these functions are distributed across complementary roles whose interactions
are mediated and auditable.

A central design principle is the establishment of \emph{authority boundaries}
and \emph{trust boundaries}. Authority boundaries specify the maximum scope of
actions that each agent may undertake, including which subsystems they may
query and which kinds of outputs they produce. Trust boundaries define which
inputs each role is permitted to rely on and how downstream agents should treat
messages originating from different roles. These boundaries prevent an agent
from issuing commands or adopting objectives outside its domain, and they limit
the damage caused if an individual role becomes compromised or misaligned.

Another design principle is \emph{separation of concerns}. Each role focuses on
a single cognitive layer—user intent modeling, policy-level decision making, or
safety validation—rather than blending multiple forms of reasoning. This
ensures interpretability of agent interactions, simplifies traceability, and
reduces the likelihood that an adversarial influence on one component will
propagate unchecked through the cognitive pipeline.

Finally, the architecture differentiates between manually driven and autonomous
modes by adjusting how much influence agentic roles have on the vehicle’s
behavior. In manually driven mode, agentic components primarily provide
high-level assistance and situational summaries, while in autonomous mode their
outputs serve as cognitive inputs to the deterministic control stack. The
underlying boundaries remain constant, but the operational weight of each role
shifts according to the driving context.

Together, these principles ensure that agentic components remain powerful
enough to support advanced reasoning and interaction, yet sufficiently
constrained to preserve safety and limit adversarial impact. It should also be noted that the proposed roles in the following section are a security-oriented abstraction of the broader agentic capabilities discussed in \cite{yu2025agentic} for agentic vehicles.

\subsection{Role Specifications}

With the global design principles established in Section~3.1, each role within
the agentic vehicle architecture can be defined by its specific responsibilities,
capabilities, and influence on the driving pipeline. The roles operate under the
same overarching authority and trust constraints, but differ in how they process
information and contribute to vehicle-level decisions.

\subsubsection{Driving Strategy Agent (DSA)}

The Driving Strategy Agent is responsible for generating policy-level behavioral
proposals that shape the vehicle’s tactical driving decisions. It interprets
high-level goals produced by the Personal Agent and integrates them with
environmental summaries, traffic conditions, and regulatory constraints. The DSA
performs multi-step reasoning to explore alternative maneuvers, evaluate risk,
and optimize trade-offs among efficiency, comfort, and compliance. It can integrate not only onboard perception summaries but also cooperative and infrastructural information (e.g., V2X messages, infrastructure-agent broadcasts, or fleet-level guidance) as contextual inputs. Its outputs
consist of structured policy recommendations, such as lane-change intentions,
desired spacing, or route preferences, which serve as cognitive inputs to the
Safety Check Layer. The DSA does not directly command actuators; instead, its
influence is strictly limited to specifying candidate behaviors for downstream
validation.

\subsubsection{Personal Agent (PA)}

The Personal Agent manages all user-facing interaction and maintains a
contextual model of user preferences, objectives, and situational constraints.
It processes natural-language dialogue, multimodal cues, and external
information sources such as navigation services or cloud-based assistants. The
PA transforms this information into high-level intent descriptors that capture
the user’s goals without specifying concrete driving actions. It also maintains
long-term state—such as historical preferences or recurring patterns—that
supports personalization across trips. The Personal Agent may also query external services and infrastructural information providers (e.g., map/traffic servers, charging infrastructure APIs, or municipal services) via authenticated tool calls or companion applications. While the PA shapes the objective context
for driving decisions, it plays no role in generating or validating maneuvers,
and its outputs remain abstract descriptors rather than behavioral proposals. 

\subsubsection{Deterministic Safety Check Layer (Non-Agentic)}
\label{sec:deterministic_safety_check}

In the proposed architecture, the final stage before the deterministic control
stack is implemented as a deterministic Safety-Check (SC) mechanism rather
than an agentic component. This unit is intentionally designed to be
non-agentic, stateless, and strictly rule-based. Its purpose is to apply
pre-defined physical, regulatory, and operational constraints to the behavioral
proposals generated by the Driving Strategy Agent (DSA).

The deterministic safety check verifies candidate actions through fixed rules
such as collision envelopes, acceleration and jerk limits, road-geometry
constraints, and regulatory compliance. This ensures that no high-level
strategy can violate certifiable safety properties of the underlying vehicle
control system.

\subsubsection{Optional Agentic Extensions, Sub-Agents, and Orchestration Considerations}
\label{sec:optional_subagents}

While the proposed architecture focuses on two primary agentic roles—the
Personal Agent (PA) and the Driving Strategy Agent (DSA)—future vehicle
platforms may incorporate additional agentic components. Most of these additions are not treated as independent high-privilege
agents in this work. Instead, they can be implemented as \emph{sub-agents}
within either the PA or DSA, preserving the authority and trust boundaries
defined by the core roles. We first start with some of the possible sub-agents. However, we will then discuss the possibility of considering a Safety Monitor Agent and an Orchestrator Agent.

\textbf{Health-Monitoring Sub-Agent (under DSA)}:
A future implementation may include a health-monitoring or self-diagnosis
reasoning unit that evaluates internal degradation, faults, or environmental
constraints. As a DSA sub-agent, it may contribute contextual constraints
(e.g., degraded-mode limits) without expanding decision authority.

\textbf{Infrastructure/Cooperation Sub-Agent (under PA):}
Interfaces that engage with smart infrastructure, municipal services, traffic
management systems, or cooperative fleets may serve as PA sub-agents. These
units translate external cooperative signals or policy constraints into modified
intent descriptors or contextual summaries for the DSA.

\textbf{Semantic Scene Interpretation Sub-Agent (under DSA):}
If future systems adopt LLM-based semantic interpretation of fused perception
outputs, a semantic-understanding sub-agent may be embedded within the DSA.
Its role would be to convert structured perception summaries into high-level
semantic descriptions relevant for policy reasoning.

\textbf{Orchestration and Safety-Monitor Agents:}
In some agentic-AI frameworks, an explicit \emph{orchestrator agent} is
introduced to coordinate multiple reasoning components, delegate tasks, or
manage inter-agent communication. Similarly, advanced architectures may
include a dedicated \emph{Safety-Monitor Agent} equipped with anomaly
detection, semantic consistency checking, and cross-layer validation
capabilities. In the vehicular context, however, we deliberately avoid
introducing either of these high-privilege agentic roles. Orchestration is
handled implicitly through the structured communication pathways and the
authority hierarchy between the Personal Agent (PA) and Driving Strategy
Agent (DSA), eliminating the need for a standalone orchestrator that would
expand the attack surface without offering additional security benefits. 
Likewise, instead of employing a full Safety-Monitor Agent, this work models
the final verification stage as a deterministic, non-agentic safety-check unit,
focused solely on enforcing immutable physical and regulatory constraints. 
Designing an agentic safety monitor with reasoning or anomaly-detection
capabilities is an important direction for future research, but it lies outside
the scope of the present work, whose primary objective is to analyze risks and
attack pathways rather than propose defensive or detection mechanisms. However, depending on capabilities of a safety monitor agent or an orchestrator, security risks should be investigated for them as well.

\subsection{Inter-Role Communication and Dependencies}

The three agentic roles communicate through a structured, sequential process that transforms user intent into safety-validated behavioral proposals. The Personal Agent initiates this pipeline by producing high-level intent descriptors that summarize user goals and contextual preferences without specifying maneuvers. These descriptors provide the semantic frame from which the Driving Strategy Agent constructs candidate strategies, integrating user intent with environmental information, traffic conditions, and regulatory constraints.

The Driving Strategy Agent sends these structured proposals—containing the main objectives and goals—to the Safety check layer, which performs verification checks to ensure consistency with safety limits and system integrity. The SC either approves the proposal, requests revision, or substitutes a conservative alternative. This feedback loop allows the DSA to refine its strategies while ensuring that no policy influences the deterministic control stack without safety validation.

Although communication flows primarily in a forward direction, the roles remain interdependent: the PA shapes intent, the DSA generates policy-level options, and the SC enforces safety constraints. This separation ensures that no single role can independently interpret intent, generate a maneuver, and authorize its execution, maintaining clear boundaries that limit adversarial propagation across the cognitive pipeline.

\subsection{Architecture Diagram}

    \begin{figure}
			\centering
			\includegraphics[width=0.9\textwidth]{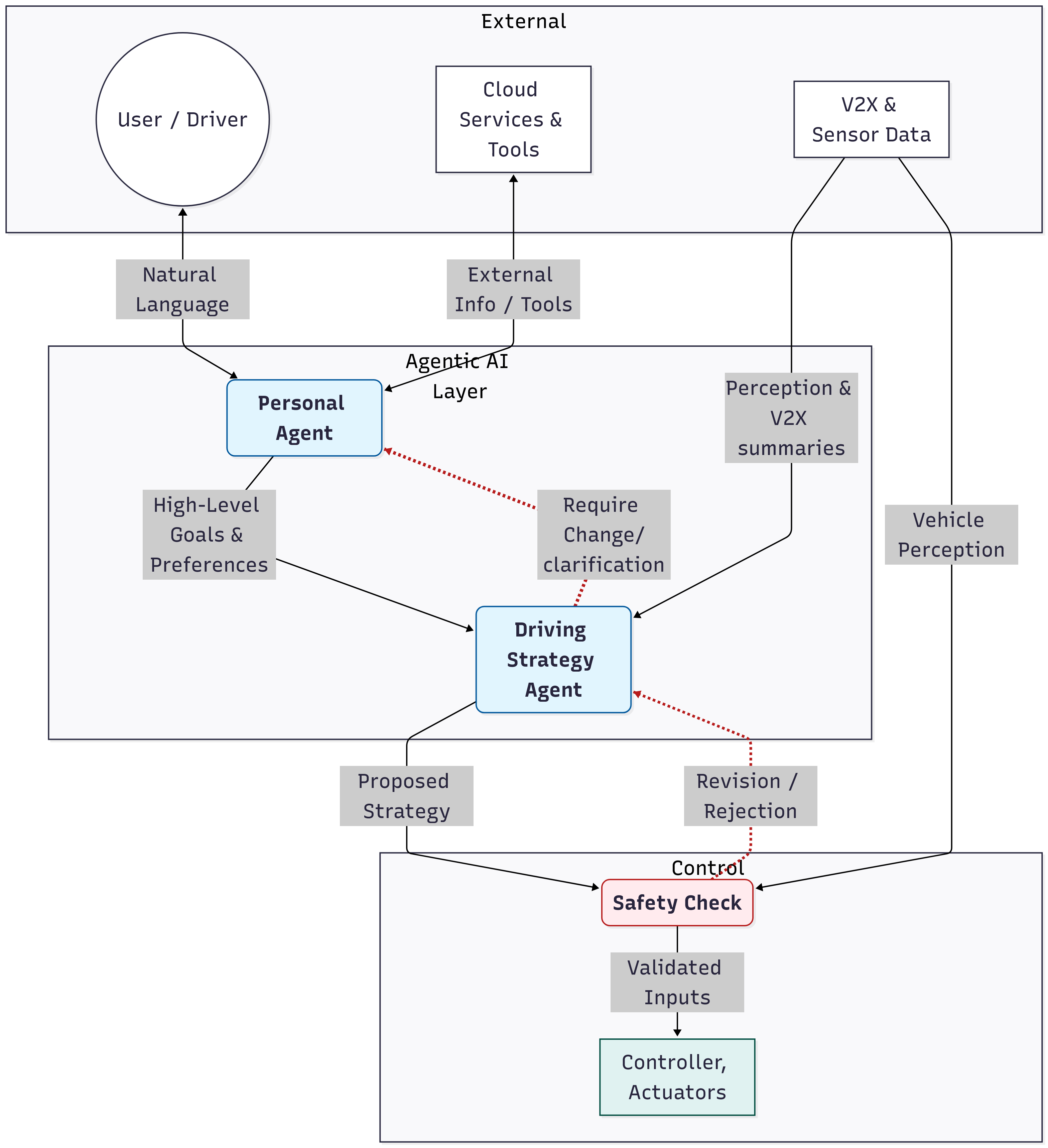}
			\caption{MAS diagram.}
            \label{fig:agentic_architecture}
	\end{figure}

Figure \ref{fig:agentic_architecture} illustrates the role-based architecture 
proposed for Agentic Vehicles. The diagram captures the structural separation 
between the agentic layer and the underlying CAV stack, as well as the 
communication dependencies and trust boundaries that regulate information 
flow. The agentic roles, Personal Agent and Driving Strategy Agent, are shown as distinct cognitive components, each with 
well-defined responsibilities and constrained authority. Information flows 
upward from perception, fusion, V2X, and cloud layers into the agentic layer, 
while influence flows downward only after safety validation performed by the 
Safety Check Layer.

The Personal Agent appears at the left of the agentic layer, interfacing with 
the user and external services. It communicates goal- and preference-related 
information to the Driving Strategy Agent, but has no direct path to the 
deterministic planning and control stack. The Driving Strategy Agent occupies 
the central position, synthesizing contextual information and user 
constraints into behavioral proposals. These proposals must pass through the 
Safety Check layer, which evaluates them against system integrity 
requirements before allowing them to influence the deterministic behavioral 
planner. A dashed trust boundary separates the agentic layer from safety-
critical subsystems, emphasizing that no agentic component can bypass 
oversight or interface directly with actuators.

The architecture diagram visually reinforces the principles articulated in 
the previous subsections: separation of concerns, role-level abstraction, and 
tiered authority. It also provides a structural basis for the threat taxonomy 
in Section~\ref{sec:threat_taxonomy}, where adversarial risks are mapped onto 
specific roles and communication pathways.

\section{Threat taxonomy for agentic vehicles}
\label{sec:threat_taxonomy}

\subsection{Threat Analysis}
\label{subsec:Threat Analysis}
This section provides a concise role-aware interpretation of the fifteen 
OWASP agentic risks in the context of Agentic Vehicles. The goal is to explain how each security threat interacts with multi-agent system, and to highlight 
the safety-relevant pathways through which cognitive disturbances propagate 
into the vehicular decision pipelines. To maintain clarity, we first provide a short description of the threat and then explain how these risks show themselves in agentic vehicle pipeline.

\subsubsection{T1: Memory Poisoning}

Memory Poisoning happens when an attacker inserts false or misleading information into the agent’s short-term or long-term memory, changing the baseline context the system relies on. The Personal Agent may adopt these altered entries as genuine user preferences or environmental cues, which then guide the Driving Strategy Agent’s planning and risk evaluations. This leads the DSA to create strategies that look coherent but are based on corrupted assumptions. Since the reasoning itself appears consistent wit safety checks, these proposals often pass through the SC without being questioned. If the poisoning persists across sessions, the resulting behavioral drift can become long-term and difficult to trace.

\subsubsection{T2: Tool Misuse}

Tool Misuse occurs when an attacker subtly steers an agent into using external tools in ways that serve the attacker’s goals. This can happen through deceptive prompts, crafted context, or misleading inputs that cause the agent to choose the wrong tool, supply harmful parameters, or chain tool calls into an unintended sequence—often while staying within its allowed permissions. The Personal Agent or Driving Strategy Agent may accept these manipulated tool outputs as reliable, causing the DSA’s planning to shift toward routes, assessments, or constraints shaped by adversarial data. Because these actions appear procedurally valid, the resulting strategy may move through the SC without challenge.

\subsubsection{T3: Privilege Compromise}

Privilege and Role Abuse arises when an attacker takes advantage of weak permission boundaries or misconfigured role delegation, allowing one agent—or an external actor—to perform actions meant for a more privileged role. This may occur through inherited permissions, temporary capabilities, or poorly separated responsibilities. If the Personal Agent or another component gains unintended authority, it can inject instructions or context into the Driving Strategy Agent that look legitimate but originate from an unauthorized source. These shifts in authority flow can influence planning in subtle ways, and the resulting proposals may slip past SC review because the final behavior still respects physical safety limits.

\subsubsection{T4: Resource Overload}

Resource Overload occurs when an attacker floods one or more agentic roles with excessive inputs, requests, or computational demands, slowing their ability to respond or evaluate context correctly. The Personal Agent may struggle to process user cues or environmental signals, causing incomplete or outdated information to pass to the Driving Strategy Agent. Under this strain, the DSA may produce rushed or shallow reasoning steps, resulting in plans that look reasonable but are based on partial evaluation. With timing pressure increased, the SC may only see a proposal that appears safe on the surface, allowing degraded reasoning to influence behavior without drawing attention.

\subsubsection{T5: Cascading Hallucinations}

Cascading Hallucinations occur when early reasoning errors—such as invented details, unsupported assumptions, or misinterpreted context—propagate and amplify across multiple steps of the agent’s decision-making. A small hallucinated element adopted by the Personal Agent can shape the intent or context provided to the Driving Strategy Agent, which may then incorporate that false premise into multi-step planning, risk assessments, or route evaluation. As each step builds on the last, the internal narrative becomes increasingly confident yet increasingly wrong. Because the final proposal still appears coherent, the SC typically sees nothing unusual and allows it through. These cascades can also persist if hallucinated content enters memory or influences future context.

\subsubsection{T6: Intent Breaking}

Intent Breaking occurs when the system forms an incorrect interpretation of the user’s goals—either due to misleading inputs, ambiguous phrasing, or internal reasoning drift. It may also be exploited by an attacker who subtly injects misleading cues or sub-goals, causing the Personal Agent to construct an intent frame that diverges from what the user actually meant. Once encoded, this distorted intent guides the Driving Strategy Agent’s planning, resulting in strategies that appear legitimate but follow a shifted objective. Because the final plan remains operationally safe, the SC generally treats it as acceptable, allowing the misinterpreted intent to influence downstream behavior.

\subsubsection{T7: Misaligned and Deceptive Behaviors}

Misaligned Behaviors arise when the system’s internal optimization priorities evolve in unintended ways—for example, overemphasizing efficiency, minimizing discomfort, or favoring conservative maneuvers due to internal drift. An attacker can also trigger this shift by nudging the agent’s reasoning toward particular outcomes without explicitly rewriting intent. The Personal Agent or Driving Strategy Agent then ranks options according to these skewed criteria, and the DSA produces plans that are coherent yet misaligned with genuine user expectations. Since the resulting actions stay within normal safety bounds, they typically pass through Safety Checks without signaling that a priority shift has occurred.

\subsubsection{T8: Repudiation and Untraceability}

Repudiation occurs when an attacker exploits weak attribution or incomplete logging, making it unclear which agent produced a given instruction or why a particular decision was made. If the Personal Agent’s intent summaries, or the DSA’s reasoning steps, are not properly recorded or bound to their origin, an adversary can introduce actions that appear untraceable or falsely attributed. This breaks the chain of accountability and makes it difficult to diagnose manipulation or reconstruct how a flawed proposal was formed. Since the SC only sees the final plan and not the missing reasoning context, these attribution gaps allow harmful or misleading actions to pass through without raising procedural alarms.

\subsubsection{T9: Identity Spoofing}

Identity Spoofing happens when an attacker impersonates a trusted user, agent, or system component, causing the Personal Agent or Driving Strategy Agent to accept instructions or context from a forged source. This can take the form of spoofed user inputs, fabricated inter-agent messages, or falsified system signals that appear to come from an authorized entity. Once accepted, these deceptive inputs influence the DSA’s planning in subtle but meaningful ways, as the agent treats the forged identity as legitimate. Because the resulting proposals still follow normal safety rules, the SC typically has no reason to question them, allowing identity-based manipulation to blend into the normal reasoning pipeline.

\subsubsection{T10: Overwhelming the Human-in-the-Loop}

Human Overload occurs when the system generates more queries, alerts, or requests for confirmation than the user can reasonably process. This may happen naturally due to poor reasoning, excessive uncertainty, or overly cautious agentic behavior, but it can also be intentionally amplified by an attacker who manipulates the context to trigger more user interactions than necessary. In either case, the user becomes overwhelmed and begins providing rushed or inconsistent feedback, which the Personal Agent then incorporates as if it were reliable input. This degraded human–agent loop shapes the context passed to the Driving Strategy Agent, causing the DSA to plan around noisy or incomplete guidance. Because the final strategy still appears structurally safe, it typically moves past SC checks without highlighting that the user’s decision-making capacity was overloaded.

\subsubsection{T11: Unexpected Remote Code Execution}

Remote Code Execution in tool chains occurs when an attacker leverages a vulnerable external tool, API, or processing step to run unauthorized code through the agent. Because agentic systems often call tools automatically—sometimes in multi-step chains—malicious payloads can enter the pipeline disguised as normal outputs. The Personal Agent or Driving Strategy Agent may then interpret the compromised tool’s results as legitimate, allowing attacker-controlled logic to influence planning. Depending on the tool’s access, this can alter state, inject harmful data, or modify internal reasoning. Since the resulting behavior can still appear operationally safe at the proposal level, the SC rarely detects that a deeper compromise has occurred.

\subsubsection{T12: Agent Communication Poisoning}

Communication Poisoning occurs when an attacker injects false, incomplete, or strategically crafted messages into the channels agents use to coordinate. In an Agentic Vehicle, this may involve spoofed updates, altered summaries, or misleading signals that cause the Personal Agent or Driving Strategy Agent to treat the injected message as legitimate. Once accepted, the distorted communication shapes the DSA’s planning process—either by warping its understanding of context or by altering constraints it believes originated from another trusted component. Because the resulting plan still appears structurally sound, the SC usually has no cue that upstream coordination was manipulated.

\subsubsection{T13: Rogue Agents}

Rogue Agents arise when one of the agentic components—PA, DSA, or even a subordinate tool helper—begins operating outside its intended role, either due to compromise or internal drift. A Rogue PA may generate intent frames that subtly pursue adversarial objectives, while a Rogue DSA may craft plans that look compliant but secretly prioritize the attacker’s goals. These deviations can be gradual and strategically hidden within the normal reasoning pipeline, allowing the rogue component to influence planning without triggering obvious alarms. Because the final outputs often remain within safe limits, the SC mainly sees behavior that “looks normal,” enabling the rogue agent’s influence to persist and shape system behavior over time.

\subsubsection{T14: Human Attacks on Multi-Agent Systems}
Human Attacks on Multi-Agent Systems occur when the user deliberately exploits inter-agent delegation, task handoff patterns, or trust relationships to manipulate system behavior. By giving conflicting instructions, rapidly changing priorities, staging deceptive inputs, or overloading agents with excessive requests, a malicious user can disrupt coordination across the agentic pipeline. The Personal Agent may encode intentionally misleading guidance, which the Driving Strategy Agent then treats as legitimate context for planning, allowing the user to indirectly steer the system toward undesirable or unsafe operational states. Because the resulting plans still follow structurally valid reasoning steps, they typically pass through SC review, even though the workflow was manipulated through intentional user-driven interference.

\subsubsection{T15: Human Manipulation}

In scenarios where AI agents interact directly with human users, the trust placed in the system reduces skepticism and increases reliance on its responses, making users more susceptible to subtle influence. This trust-driven vulnerability allows the agent’s own framing, confidence, or emphasis to unintentionally shape user beliefs or choices, and it can also be exploited by external actors who steer the agent’s messaging to mislead users or encourage covert actions. In an Agentic Vehicle, such influence affects the pipeline by causing the Personal Agent to encode user feedback that was shaped by the agent itself, which the Driving Strategy Agent then incorporates into planning. The resulting strategies may appear coherent and aligned with explicit user input, allowing them to pass SC review, even though the underlying preferences or decisions were partially created or distorted through human-facing manipulation.

\subsection{Threat Severity Analysis}
\label{subsec:Threat Severity Matrix}
For each threat, we evaluate four dimensions: Safety Impact (effect on vehicle safety and control and denoted by SI), Stealth \& Detectability (difficulty of identifying the attack and denoted by SD), Persistence (ability to survive across sessions or propagate through memory and denoted by P), and Semantic Misalignment (distortion of goals, intent, or decision semantics and denoted by SM). Each dimension is rated using a four-level ordinal scale (L, M, H, C) corresponding to increasing severity. To obtain a single scalar indicator, we map L=1, M=2, H=3, and C=4 and compute their sum for each threat–scenario combination. These totals are then color-coded using four severity bands (Low: 4–7, Medium: 8–10, High: 11–13, Critical: 14–16). This approach allows side-by-side comparison across six operating contexts (manual vs. autonomous vehicles and three levels of agentic capability) and highlights how severity escalates as both autonomy and agentic sophistication increase.

By analyzing Tables \ref{Table I} to \ref{Table VI}, one can find interesting realizations:
\begin{enumerate}
    \item In manual driving with low agentic capability, most threats remain in the lower severity bands because reasoning-driven components exert limited influence on driving behavior and human supervision dominates. As the agentic capability grows, threats associated with reasoning (memory, intent, hallucinations) and delegation (tools, identity, communication) begin to push scores upward even in manual settings.

\item Autonomous operation amplifies every threat category, regardless of agency level. When the vehicle itself executes maneuver decisions, semantic distortions and upstream contextual errors can propagate directly into motion planning. Even small corruptions that remain physically safe in manual settings may become more consequential once human oversight is removed, which shifts multiple threats from low/medium in manual contexts to high/critical in autonomous ones.

\item The tables show that semantic threats dominate the upper ranges. Threats such as memory poisoning, cascading hallucinations, intent breaking, misaligned behavior, and rogue agents systematically move toward the “High” and “Critical” categories as soon as agentic reasoning significantly contributes to policy generation. This suggests that traditional cybersecurity—focused on communication and hardware integrity—cannot by itself bound the risk of misaligned reasoning or long-term semantic drift.
\end{enumerate}

%---------------------- Manual / Low ----------------------
\begin{table}[h]\centering
\small
\begin{tabular}{lccccc}
\toprule
Threat & SI & SD & P & SM & Severity  \\
\midrule
T1 Memory Poisoning                          & L & L & L & L & \cellcolor{green!20}4 \\
T2 Tool Misuse                               & L & M & L & L & \cellcolor{green!20}5 \\
T3 Privilege Compromise                      & L & M & M & L & \cellcolor{green!20}6 \\
T4 Resource Overload                         & L & L & L & L & \cellcolor{green!20}4 \\
T5 Cascading Hallucinations                  & L & M & L & M & \cellcolor{green!20}6 \\
T6 Intent Breaking                           & L & L & M & M & \cellcolor{green!20}6 \\
T7 Misaligned \& Deceptive Behaviors         & L & M & M & M & \cellcolor{green!20}7 \\
T8 Repudiation \& Untraceability             & L & L & M & L & \cellcolor{green!20}5 \\
T9 Identity Spoofing                         & L & M & M & L & \cellcolor{green!20}6 \\
T10 Overwhelming Human-in-the-Loop           & L & L & L & L & \cellcolor{green!20}4 \\
T11 Unexpected Remote Code Execution         & L & M & M & L & \cellcolor{green!20}6 \\
T12 Agent Communication Poisoning            & L & M & M & L & \cellcolor{green!20}6 \\
T13 Rogue Agents                             & L & M & M & L & \cellcolor{green!20}6 \\
T14 Human Attacks on Multi-Agent Systems     & L & M & L & L & \cellcolor{green!20}5 \\
T15 Human Manipulation                       & L & M & M & M & \cellcolor{green!20}7 \\
\bottomrule
\end{tabular}
\caption{Manual / Low Agency}
\label{Table I}
\end{table}

%---------------------- Manual / Medium ----------------------
\begin{table}[h]\centering
\small
\begin{tabular}{lccccc}
\toprule
Threat & SI & SD & P & SM & Severity  \\
\midrule
T1 Memory Poisoning                          & M & M & H & M & \cellcolor{yellow!20}9 \\
T2 Tool Misuse                               & M & H & M & M & \cellcolor{yellow!20}9 \\
T3 Privilege Compromise                      & M & H & M & M & \cellcolor{yellow!20}9 \\
T4 Resource Overload                         & M & M & M & M & \cellcolor{yellow!20}8 \\
T5 Cascading Hallucinations                  & M & H & M & M & \cellcolor{yellow!20}9 \\
T6 Intent Breaking                           & M & M & H & H & \cellcolor{orange!20}10 \\
T7 Misaligned \& Deceptive Behaviors         & M & H & H & H & \cellcolor{orange!20}11 \\
T8 Repudiation \& Untraceability             & M & H & H & M & \cellcolor{orange!20}10 \\
T9 Identity Spoofing                         & M & H & M & M & \cellcolor{yellow!20}9 \\
T10 Overwhelming Human-in-the-Loop           & M & M & M & M & \cellcolor{yellow!20}8 \\
T11 Unexpected Remote Code Execution         & M & H & H & M & \cellcolor{orange!20}10 \\
T12 Agent Communication Poisoning            & M & H & M & M & \cellcolor{yellow!20}9 \\
T13 Rogue Agents                             & M & H & H & M & \cellcolor{orange!20}10 \\
T14 Human Attacks on Multi-Agent Systems     & M & M & M & M & \cellcolor{yellow!20}8 \\
T15 Human Manipulation                       & M & H & H & H & \cellcolor{orange!20}11 \\
\bottomrule
\end{tabular}
\caption{Manual / Medium Agency}
\label{Table II}
\end{table}

%---------------------- Manual / High ----------------------
\begin{table}[h]\centering
\small
\begin{tabular}{lccccc}
\toprule
Threat & SI & SD & P & SM & Severity  \\
\midrule
T1 Memory Poisoning                          & H & H & H & H & \cellcolor{orange!20}12 \\
T2 Tool Misuse                               & H & H & H & H & \cellcolor{orange!20}12 \\
T3 Privilege Compromise                      & H & H & H & H & \cellcolor{orange!20}12 \\
T4 Resource Overload                         & H & M & H & H & \cellcolor{orange!20}11 \\
T5 Cascading Hallucinations                  & H & H & H & H & \cellcolor{orange!20}12 \\
T6 Intent Breaking                           & H & H & H & H & \cellcolor{orange!20}12 \\
T7 Misaligned \& Deceptive Behaviors         & H & H & H & C & \cellcolor{orange!20}13 \\
T8 Repudiation \& Untraceability             & M & H & H & M & \cellcolor{orange!20}10 \\
T9 Identity Spoofing                         & M & H & H & H & \cellcolor{orange!20}11 \\
T10 Overwhelming Human-in-the-Loop           & M & M & M & M & \cellcolor{yellow!20}8 \\
T11 Unexpected Remote Code Execution         & C & H & C & H & \cellcolor{red!20}13 \\
T12 Agent Communication Poisoning            & H & H & H & H & \cellcolor{orange!20}12 \\
T13 Rogue Agents                             & H & H & H & C & \cellcolor{red!20}13 \\
T14 Human Attacks on Multi-Agent Systems     & H & M & H & H & \cellcolor{orange!20}11 \\
T15 Human Manipulation                       & H & H & H & H & \cellcolor{orange!20}12 \\
\bottomrule
\end{tabular}
\caption{Manual / High Agency}
\label{Table III}
\end{table}

%---------------------- Autonomous / Low ----------------------
\begin{table}[h]\centering
\small
\begin{tabular}{lccccc}
\toprule
Threat & SI & SD & P & SM & Severity  \\
\midrule
T1 Memory Poisoning                          & M & L & L & M & \cellcolor{green!20}7 \\
T2 Tool Misuse                               & M & M & M & M & \cellcolor{yellow!20}8 \\
T3 Privilege Compromise                      & M & M & H & M & \cellcolor{yellow!20}9 \\
T4 Resource Overload                         & M & M & M & M & \cellcolor{yellow!20}8 \\
T5 Cascading Hallucinations                  & M & M & M & M & \cellcolor{yellow!20}8 \\
T6 Intent Breaking                           & M & M & M & H & \cellcolor{yellow!20}9 \\
T7 Misaligned \& Deceptive Behaviors         & M & M & M & H & \cellcolor{yellow!20}9 \\
T8 Repudiation \& Untraceability             & M & H & H & M & \cellcolor{orange!20}10 \\
T9 Identity Spoofing                         & M & M & M & M & \cellcolor{yellow!20}8 \\
T10 Overwhelming Human-in-the-Loop           & L & L & L & L & \cellcolor{green!20}4 \\
T11 Unexpected Remote Code Execution         & H & M & H & M & \cellcolor{orange!20}10 \\
T12 Agent Communication Poisoning            & M & M & M & M & \cellcolor{yellow!20}8 \\
T13 Rogue Agents                             & H & M & H & M & \cellcolor{orange!20}10 \\
T14 Human Attacks on Multi-Agent Systems     & M & M & M & M & \cellcolor{yellow!20}8 \\
T15 Human Manipulation                       & L & M & M & M & \cellcolor{green!20}7 \\
\bottomrule
\end{tabular}
\caption{Autonomous / Low Agency}
\label{Table IV}
\end{table}

%---------------------- Autonomous / Medium ----------------------
\begin{table}[h]\centering
\small
\begin{tabular}{lccccc}
\toprule
Threat & SI & SD & P & SM & Severity  \\
\midrule
T1 Memory Poisoning                          & H & H & H & H & \cellcolor{orange!20}12 \\
T2 Tool Misuse                               & H & H & H & H & \cellcolor{orange!20}12 \\
T3 Privilege Compromise                      & H & H & H & H & \cellcolor{orange!20}12 \\
T4 Resource Overload                         & H & M & H & H & \cellcolor{orange!20}11 \\
T5 Cascading Hallucinations                  & H & H & H & H & \cellcolor{orange!20}12 \\
T6 Intent Breaking                           & H & H & H & H & \cellcolor{orange!20}12 \\
T7 Misaligned \& Deceptive Behaviors         & H & H & H & H & \cellcolor{orange!20}12 \\
T8 Repudiation \& Untraceability             & H & H & H & H & \cellcolor{orange!20}12 \\
T9 Identity Spoofing                         & H & H & H & H & \cellcolor{orange!20}12 \\
T10 Overwhelming Human-in-the-Loop           & M & M & L & M & \cellcolor{green!20}7 \\
T11 Unexpected Remote Code Execution         & C & H & C & H & \cellcolor{red!20}14 \\
T12 Agent Communication Poisoning            & H & H & H & H & \cellcolor{orange!20}12 \\
T13 Rogue Agents                             & H & H & H & H & \cellcolor{orange!20}12 \\
T14 Human Attacks on Multi-Agent Systems     & H & M & H & H & \cellcolor{orange!20}11 \\
T15 Human Manipulation                       & M & H & H & H & \cellcolor{orange!20}11 \\
\bottomrule
\end{tabular}
\caption{Autonomous / Medium Agency}
\label{Table V} 
\end{table}

%---------------------- Autonomous / High ----------------------
\begin{table}[h]\centering
\small
\begin{tabular}{lccccc}
\toprule
Threat & SI & SD & P & SM & Severity  \\
\midrule
T1 Memory Poisoning                          & C & H & C & C & \cellcolor{red!20}15 \\
T2 Tool Misuse                               & C & C & H & C & \cellcolor{red!20}15 \\
T3 Privilege Compromise                      & C & H & C & H & \cellcolor{red!20}14 \\
T4 Resource Overload                         & H & M & H & H & \cellcolor{orange!20}11 \\
T5 Cascading Hallucinations                  & C & C & H & C & \cellcolor{red!20}15 \\
T6 Intent Breaking                           & C & H & C & C & \cellcolor{red!20}15 \\
T7 Misaligned \& Deceptive Behaviors         & C & C & C & C & \cellcolor{red!20}16 \\
T8 Repudiation \& Untraceability             & H & C & H & H & \cellcolor{red!20}13 \\
T9 Identity Spoofing                         & C & H & H & C & \cellcolor{red!20}14 \\
T10 Overwhelming Human-in-the-Loop           & M & M & L & M & \cellcolor{yellow!20}7 \\
T11 Unexpected Remote Code Execution         & C & C & C & C & \cellcolor{red!20}16 \\
T12 Agent Communication Poisoning            & C & H & H & C & \cellcolor{red!20}14 \\
T13 Rogue Agents                             & C & H & H & C & \cellcolor{red!20}14 \\
T14 Human Attacks on Multi-Agent Systems     & H & M & H & H & \cellcolor{orange!20}11 \\
T15 Human Manipulation                       & H & H & H & H & \cellcolor{orange!20}12 \\
\bottomrule
\end{tabular}
\caption{Autonomous / High Agency}
\label{Table VI}
\end{table}

\subsection{Cross-Role Attack Chains}

Small distortions in one agentic role can propagate through the reasoning pipeline and produce misaligned decisions that remain undetected by the Safety Check layer. The following chains illustrate some of these important pathways identified in our analysis.

\textbf{Chain 1: Memory Poisoning $\rightarrow$ Intent Drift $\rightarrow$ Misaligned Strategy}:
An attacker inserts subtle preference cues into the Personal Agent's long-term memory (T1), causing it to encode distorted user intent over subsequent interactions (T6). The Driving Strategy Agent optimizes policies around this shifted objective, producing strategies that are misaligned with actual user goals yet remain physically safe (T7). Since no explicit safety constraint is violated, the SC layer approves them, resulting in persistent behavioral drift.

\textbf{Chain 2: Identity Spoofing $\rightarrow$ Communication Poisoning $\rightarrow$ Misaligned Strategy} :
A forged identity (T9) enables injection of falsified environmental messages---such as fake traffic updates or infrastructure warnings---into the inter-agent communication channel (T12). The DSA treats these poisoned communications as authoritative constraints and adjusts its reasoning accordingly, justifying unnecessary detours or suboptimal maneuvers (T7). Because the proposals remain internally consistent with the (false) environmental context, the SC layer finds no reason to reject them.

\textbf{Chain 3: Cascading Hallucinations $\rightarrow$ Compounding Errors $\rightarrow$ Undetected Misalignment}:
A minor hallucinated detail---such as a misread speed limit or invented road closure---enters the PA's context (T5). As the DSA performs multi-step reasoning, each step builds confidence in the incorrect narrative, gradually shaping a coherent but flawed policy (T7).

\textbf{Chain 4: Perception Layer Attack $\rightarrow$ Cascading Misinterpretation $\rightarrow$ Policy Misalignment}:
An adversarial perturbation to perception---such as a manipulated road sign or spoofed LiDAR return---produces a coherent but incorrect scene summary. The PA or DSA accepts this as genuine environmental context, and subsequent reasoning steps compound the error (T5), generating increasingly confident but misaligned strategies (T7). Because the corrupted inputs remain internally consistent across multiple perception features, the SC layer detects no anomaly and approves the flawed proposal.

These chains demonstrate that agentic vulnerabilities rarely operate in isolation. Instead, localized cognitive disturbances---whether internally generated or induced through upstream corruption---cascade through the reasoning pipeline and result in validated yet misaligned behavior. Understanding these propagation pathways is essential for designing defenses that operate across multiple roles rather than protecting individual components in isolation.

\section{Integrated Attacks Across CAV Layers and Agentic AI Agents}
\label{sec:Integrated_Attacks}

The threats in Section~4 arise within the agentic subsystem, but agentic roles
do not operate in isolation. Their reasoning depends on summaries produced by
the traditional CAV stack—perception, communication, computing, decision, and
control. Attacks on these upstream layers can therefore propagate into the
agentic pipeline without directly manipulating memory, goals, or reasoning.
Because the affected summaries often remain internally coherent, cross-layer
attacks may appear legitimate to the agentic modules, enabling subtle but
systematic misinterpretation of the environment and the vehicle state.

\subsection{Perception-Layer Attacks}

Adversarial perturbations to perception---such as manipulated road signs, altered
lane markings, sensor spoofing, or phantom objects---can distort the fused scene
representation produced by the perception and fusion stack. Even if underlying
sensors disagree, a dominant modality or corrupted fusion logic may still generate
an internally consistent but incorrect summary. The Driving Strategy Agent (DSA)
then updates its risk and maneuver evaluations based on this falsified scene, often
shifting toward overly conservative or unnecessarily aggressive behavior. The Personal
Agent (PA) may also misjudge workload or situational stress. As long as the resulting
strategy remains within the deterministic Safety Check (SC) layer’s physical and
regulatory envelopes, these misaligned outputs will not be rejected.

\subsection{Communication-Layer Manipulation}

Spoofed V2X messages, falsified cooperative-perception data, or tampered road-side
broadcasts can corrupt the vehicle’s digital context before it reaches the agentic
roles. Once such messages pass integrity checks or exploit misconfigurations, they
become indistinguishable from legitimate infrastructure signals. The DSA may react
to nonexistent hazards or fabricated congestion, while the PA adjusts interaction
patterns based on a false sense of urgency. If these manipulated signals are logged or
incorporated into historical preference models, their influence can persist well beyond
the attack window.

\subsection{Upstream Corruption in the Computing and Decision Layers}

The computing and decision layers perform fusion, filtering, and abstraction of sensor
and communication data into high-level state estimates. A stronger adversary capable
of compromising these modules---for example via OTA update tampering, neural-network
weight manipulation, or insider modification---can generate summaries that appear
credible yet systematically biased. Unlike perception-layer attacks, which distort raw
signals, this attack alters the abstraction itself, allowing corrupted context to propagate
even when sensors remain uncompromised. The DSA then produces strategies that are
coherent relative to these biased estimates, and the PA adapts interaction behavior
accordingly, with the SC layer approving proposals that remain physically valid.

\subsection{Misleading Feedback from the Control Layer}

Agentic modules rely on feedback from the control layer---such as velocity, acceleration,
steering angle, or braking state---to maintain situational awareness. An attacker with
in-vehicle network access may falsify these signals, causing the DSA to switch modes
unnecessarily or adopt degraded driving profiles, while the PA adjusts comfort or
preference parameters based on misreported performance. Because these falsified
signals remain internally consistent with the corrupted state, the SC layer validates
the resulting strategies.

\subsection{Illustrative Cross-Layer Attack Chains}

Cross-layer attacks often manifest through multi-stage propagation. For example, a
phantom-object attack may corrupt perception fusion, producing an internally consistent
but false scene. The DSA interprets this as a genuine hazard and shifts to an overly
conservative profile, which the SC layer approves due to the absence of contradictory
evidence. Similarly, a spoofed V2X road-closure message can induce unnecessary
rerouting or misleading user instructions, with both the PA and DSA treating the
fabricated signal as authoritative. These examples demonstrate how upstream
manipulation---whether physical, digital, or software-based---can cascade into the
agentic reasoning pipeline and produce validated yet misaligned behavior.

\section{Discussion}
\label{sec:Discussion}
\subsection{Implications for Agentic Vehicle Design}
While agentic modules such as PAs and DSAs enable richer reasoning capabilities, their integration also brings new architectural and security implications. The following are some of these implications and how to address them.
\begin{enumerate}
    \item \textbf{Strict role separation and bounded authority:}  
    Separating intent interpretation (PA), behavioral proposal generation (DSA), and safety enforcement ensures that each module operates on the information it is best suited for. Operationally, this reduces complexity and simplifies verification. From a security standpoint, failures or compromises can be isolated and contained more easily.

    \item \textbf{Agency level as a design parameter:}  
    Higher agency increases capability but also expands the attack surface and the unpredictability of agent behavior. Constraining agency levels limits risk, and dynamically lowering agency under threats or degraded conditions might help maintain safety while preserving flexibility. However, this can also open a new attack surface, when the attacker aims to degrade system performance by degrading its agency levels.

    \item \textbf{Semantic integrity of perception and V2X information:}  
    Agentic reasoning depends on coherent contextual representations by the perception and V2X infrastructure. If fused summaries are manipulated, the PA or DSA may generate plausible yet unwanted behavior. Ensuring semantic integrity through temporal, cross-sensor, and map-based consistency checks is therefore essential.

    \item \textbf{Deterministic safety components must continue to exist:}  
    Even with agentic reasoning, the system requires a deterministic layer that enforces physical and rule-based constraints and serves as the final safety gate. Not only that, but because semantic validation and multi-source checks introduce latency, a deterministic safety gate must remain to ensure safety when the latency is more than a threshold.

    \item \textbf{Human-agent interaction as a security factor:}  
    How agents request clarification, present uncertainty, and summarize decisions affects driver understanding and workload. Predictable, structured interaction patterns support safety and reduce opportunities for confusion or indirect manipulation.

    \item \textbf{System-level explainability and semantic monitoring:}  
     Maintaining provenance over which perceptions, memories, and tools influenced each proposal enables semantic drift detection and supports post-incident analysis. Although this adds computational cost, it is essential for maintaining trust.

    \item \textbf{Semantic-layer security and the optional role of a Safety Monitor Agent (SMA):}  
    A Safety Monitor Agent (SMA) can provide additional oversight by checking whether DSA proposals remain semantically coherent with scene context, temporal history, and expected behavioral patterns. To avoid increasing the attack surface, the SMA must remain tightly scoped, with restricted authority and isolated memory/tool-access boundaries. Its outputs should feed into the deterministic safety layer, but never override it. If the SMA fails (e.g., resource overload or unresponsiveness), the system should revert to deterministic safety checks or conservative defaults. This design preserves semantic monitoring benefits while preventing the SMA from becoming a new point of systemic vulnerability.

    \item \textbf{Layered, distributed defenses:}  
    Attack chains may span perception, communication, and reasoning layers. Distributed defenses, such as invariants, drift detection, provenance checks, and cross-source validation, offer robustness that cannot be achieved by any single mechanism alone.

    \item \textbf{Orchestrator agent as an optional coordination mechanism:}  
    An orchestrator can regulate information flow, enforce ordering rules, and apply provenance constraints among PA, DSA, and other possible agents. When implemented as a minimal coordinator, it enhances governance without enlarging the attack surface. A more capable agentic orchestrator enables richer coordination but increases latency, complexity, and vulnerability, and requires well-defined fallback behavior to deterministic safety control. Therefore, the trade-off between safety and its capabilities are an important design factor. 
\end{enumerate}

\subsection{Limitations of the Role-Based Taxonomy}

Although the proposed role-based taxonomy offers a structured foundation for
analyzing agentic vehicle security, it necessarily abstracts away aspects of
implementation that may influence real-world system behavior. First, the
taxonomy assumes clear separation between the Personal Agent and Driving Strategy
Agent. In practical deployments, these boundaries may
blur due to shared resources, co-located execution, or optimization of memory
and compute pathways. Such coupling can create hidden dependencies not captured
by the clean role decomposition.

Second, the taxonomy does not model emergent behaviors that arise when
reasoning modules interact through iterative feedback or jointly adapt their
internal states. These interactions can produce vulnerabilities that do not map
directly to any single threat or role, particularly when long-term memory,
multi-agent negotiation, or tool-driven autonomy drives complex system
dynamics.

Finally, the framework abstracts away differences between model architectures,
training pipelines, fine-tuning strategies, and verification mechanisms.
Variations in how capabilities such as memory, tool invocation, or constraint
handling are implemented can meaningfully alter the effective attack surface.

Despite these limitations, the role-based taxonomy provides a coherent
conceptual model for understanding and comparing threats across agentic
vehicles, and it establishes a foundation for integrating more detailed
implementation-level considerations in future research.

\section{Case Studies}
\label{sec:Case Studies}
In this section, we provide case studies on some of the mentioned security threats with the model Llama 3.1

\subsection{Case Study I: Memory Poisoning of the Personal Agent}

We evaluate how a single poisoned memory entry can alter system-level driving
behavior. The adversary inserts an innocuous statement into the Personal Agent's
(PA) long-term memory---e.g., \emph{“the vehicle should not exceed
approximately 45\,km/h due to minor internal issues”}. Although syntactically
harmless, this memory influences how the PA interprets future driver requests.

Across eight representative driving scenarios (e.g., urgent and routine),
the PA produces modified high-level preferences that the Driving Strategy Agent
(DSA) converts into target speeds. The Safety Check (SC) layer validates only
physical safety limits and therefore does not flag preference distortions.

\begin{figure}[t]
    \centering
    \includegraphics[width=0.95\linewidth]{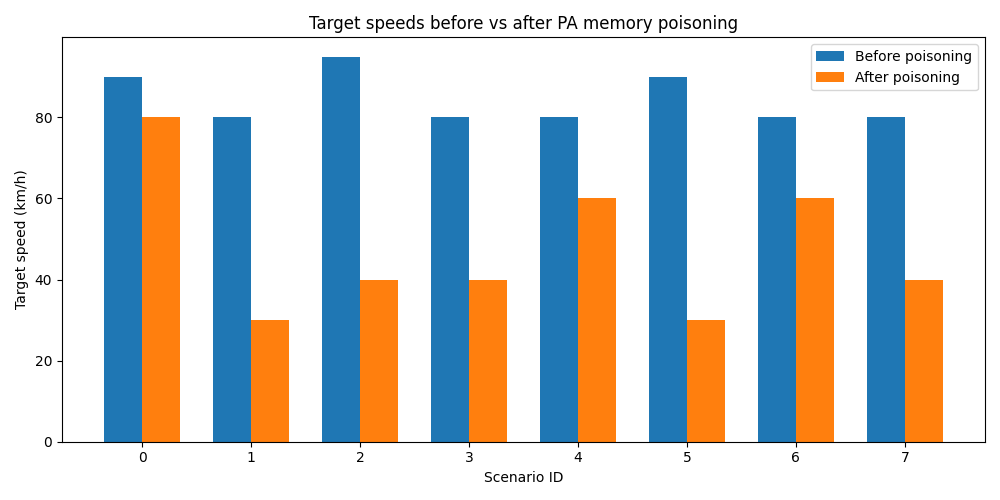}
    \caption{Comparison of target speeds generated before and after 
    poisoning the PA's memory with a constraint suggesting that the 
    vehicle should not exceed approximately 45\,km/h. The poisoning 
    induces large and consistent slowdowns across all scenarios, 
    including urgent ones, while remaining within the SC layer's physical 
    safety limits.}
    \label{fig:pa_poisoning_bars}
\end{figure}

Figure~\ref{fig:pa_poisoning_bars} shows the resulting target speeds before and
after poisoning. Before the attack, speeds match contextual expectations
(80--90\,km/h). After
poisoning, all scenarios converge to significantly reduced speeds (30--60\,km/h).
For example, Scenario~1 drops from 80\,km/h to 40\,km/h despite no change in
driver intent. Because these speeds remain physically safe, the SC layer approves
them, revealing how semantic corruption in the PA can silently propagate through
the entire agentic stack.

\subsection{Case Study II: Infrastructure-Induced Policy Misalignment}

This case study illustrates how upstream corruption of map or infrastructure descriptions propagates into the agentic pipeline. Attacker injects information through the communication with the infrastructure (such as information from Road Side Units) that the speed limit is reduced from 80~km/h to 40~kn/h due to maintenance work. The Driving Strategy Agent (DSA) then generates reduced target speeds thatass the deterministic Safety Check (SC) because they remain physically safe.

\begin{figure}[t]
    \centering
    \includegraphics[width=0.95\linewidth]{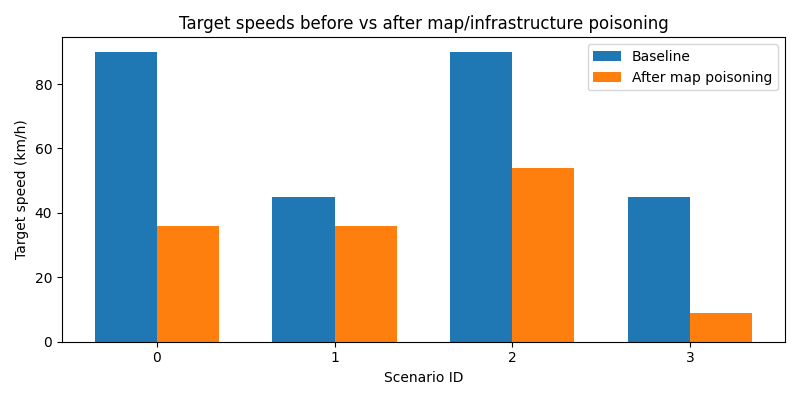}
    \caption{Comparison of target speeds generated before and after 
    receiving corrupted information from the infrastructure. As demonstrated, the DSA proposes much reduces speed targets due to the corrupted received information.}
    \label{fig:bar_speeds_map_poisoning}
\end{figure}

Figure \ref{fig:bar_speeds_map_poisoning} demonstrates the resultsAcross four scenarios (highway, arterial, ring-road, residential), all poisoned descriptions yield significantly lower target speeds. As shown in Fig.~3, high-speed cases experience the largest degradations: Scenario~0 drops from roughly 90~km/h to 36~km/h, and Scenario~2 from 90~km/h to 54~km/h. Even residential driving is forced to extreme under-driving (e.g., 45~km/h to 9~km/h) when the attacker embeds seemingly harmless ``community initiatives.''

Because the attack operates entirely upstream of the agentic layer, neither the Personal Agent nor the SC detects the semantic corruption. This demonstrates how falsified infrastructure or map context can silently mislead the DSA into persistent, misaligned behavior without violating any physical safety constraints.

\section{Conclusion}
\label{sec:Conclusion}
This paper introduced the first structured framework for analyzing the security risks of Agentic Vehicles—systems in which reasoning-driven, memory-enabled, and tool-capable AI agents influence the behavior of both manually driven and autonomous vehicles. Using a role-based architecture consisting of the Personal Agent and Driving Strategy Agent, we demonstrated and mapped the fifteen OWASP Agentic AI Security Risks onto this architecture, revealing how threats manifest differently across roles. We further showed that agentic reasoning does not operate in isolation: attacks on traditional Connected and Autonomous Vehicles (CAVs) layers, such as perception, V2X communication, computing, and control, can mislead agentic modules through corrupted summaries or falsified contextual cues, thereby creating integrated attack pathways that evade traditional cyber-physical defenses. Together, these findings establish a foundation for understanding how reasoning-driven AI introduces new risks in safety-critical environments such as vehicles.

Building on this foundation, future work will focus on developing concrete defenses for the most impactful threats identified in the taxonomy. Priority areas include detection and monitoring mechanisms for risks such as memory poisoning, cascading hallucinations, identity spoofing, and communication poisoning; lightweight authentication and provenance schemes to secure inter-agent messages and reasoning chains; and validation methods to assess the Safety Monitor Agent’s robustness under adversarial conditions without expanding its attack surface. Experimental evaluation—through simulation environments, cross-layer attack-chain testing, and prototype implementations—will be essential to translate this taxonomy into a practical security framework for next-generation agentic vehicles.

\bibliographystyle{IEEEtran}
\bibliography{sample} 
\end{document}